\documentclass[preprint,12pt]{elsarticle}

\makeatletter
\def\ps@pprintTitle{%
 \let\@oddhead\@empty
 \let\@evenhead\@empty
 \def\@oddfoot{}%
 \let\@evenfoot\@oddfoot}
\makeatother

\usepackage{amssymb}

\usepackage{amsmath}%
\newdefinition{definition}{Definition}

\newproof{pf}{Proof}
\usepackage{booktabs,tabu}
\usepackage{algorithm2e}
\SetKwInput{KwInput}{Input}
\SetKwInput{KwOutput}{Output}
\usepackage{float}
\usepackage{listings}
\usepackage[svgnames]{xcolor}
\begin{document}

\begin{frontmatter}



\title{A Limitation of V-Matrix based Methods}



\author[label1]{Niharika Gauraha}
\ead{niharika.gauraha@farmbio.uu.se}
\author[label2]{Akshay Chaturvedi}
\ead{akshay91.isi@gmail.com}

\address[label1]{Department of Pharmaceutical Biosciences \\
       Uppsala University\\
       Uppsala, Sweden}

\address[label2]{Computer Vision and Pattern Recognition Unit \\
       Indian Statistical Institute \\
       Kolkata, India}

\begin{abstract}
To estimate the conditional probability functions based on the direct problem setting, V-matrix based method was proposed. We construct V-matrix based constrained quadratic programming problems for which the inequality constraints are inconsistent. In particular, we would like to present that the constrained quadratic optimization problem for conditional probability estimation using V-matrix method may not have a consistent solution always.

\end{abstract}

\end{frontmatter}

\section{A limitation of V-matrix based (Direct) Method of Solving Conditional Probability Function}
V-Matrix method of estimation of conditional probability function was defined in \cite{vapnik2015v}. Here we present a limitation of  the same.
We mostly follow the notations in \cite{vapnik2015v}. 
We consider estimation of the following form of conditional probability function in pattern recognition problems:
\begin{align}
p(y = 1 \mid x) = \frac{p(x, y=1)}{p(x)}, \quad p(x)>0, \label{eq:conditionalProb}
\end{align}
from 
a training dataset of  ${\displaystyle \ell}$ points 
\begin{align*}
({x}_{1},y_{1}),\,\ldots ,\,({x}_{\ell},y_{\ell}),
\end{align*}
where $y_i \in \{ 0, 1\}$ and $x_i \in R^n$. Here we assume that the training data points are IID and generated according to an unknown probability measure $p(x,y) = p(y \mid x) p(x)$. 
In particular, we are looking for a conditional probability function in Reproducing Kernel Hilber Space (RKHS), in the form
\begin{align*}
	f(x) &= \sum_i \alpha_i K(x_i, x) = A^T\mathcal{K}(x),
\end{align*}
by solving the following optimization problem
\begin{align}
	\mathop{minimize}_{A}\; & (Y - KA)^TV 	(Y - KA) + \gamma A ^T K A,\\
	\text{subject to}\; & \textbf{0} \leq A^T \mathcal{K}(x_i) \leq \textbf{1}, \quad i = 1, \ldots, \ell\\
	\frac{1}{\ell}\sum_i  A^T  \kappa(x_i) &= \sum_i y_i / \ell = p_1,
\end{align}
where $K$ is the positive semidefinite kernel matrix, $p_1$ is the frequency of class $y=1$, estimated from the training data, $A = (\alpha_1, \ldots \alpha_\ell) $ are unknown weights to be estimated (optimization variables).
and \textbf{0} and \textbf{1} are $\ell$-dimensional vectors of zeros and ones respectively.

The above quadratic programming problem for estimation of conditional probability  can be simplified as
\begin{gather}
	\mathop{minimize}_{A}\;  A^T 	(KVK + \gamma K)A - 2 A^T KVY, \label{eq:qp_phi} \\
	\text{subject to}\;  KA \leq \textbf{1}, \label{eq:qp_phi_const1}\\
	-KA \leq \textbf{0},\text{ and } \\
	\frac{1}{\ell} A^T K \textbf{1} = p_1, \label{eq:qp_phi_const2}
\end{gather}
The kernel matrix $K$   appears as  constraint matrices in the above inequality and equality constraints. 
Since the kernel matrices are known to be positive semi-definite,
the constrains matrices for the above Quadratic Programming (QP) problem may not have full rank and there is no guarantee that the constraints will be consistent,
hence no solution could be found. 

\section{Illustration with XOR Example}
In this section, XOR classification problem is considered to show that the 
V-matrix method of conditional probability estimation (we call it v-SVM method),  fails if the constraint matrix does not have full rank.
Consider the following four points, $X$, in two dimensional feature space and their corresponding class labels $Y$. 
\begin{align*}
\begin{array}{cc}
{X = \left[ \begin{array}{cccc}
0 & 0\\
1 & 1\\
0 & 1 \\
1 & 0 \\
\end{array} \right]} 
& 
{Y = \begin{array}{c}
0 \\
0 \\
1 \\
1
\end{array}}
\end{array}
\end{align*}
The kernel matrix produced by INK-spline and RBF kernel function is given in the following.
\begin{align*}
\begin{array}{cc}
{K_{spline} = \left[ \begin{array}{cccc}
0 & 0 & 0 & 0\\
0 & 2& 1 & 1\\
0 & 1 &1 & 0\\
0 & 1 &0 & 1\\
\end{array} \right]} 
& 
{K_{rbf} = \left[ \begin{array}{cccc}
0 & 0.37 & 0.61 & 0.61\\
0.27 & 1& .61 & .61\\
.61 & .61 &1 & 0.37\\
.61 & .61 &0.37 & 1\\
\end{array} \right]} 
\end{array}
\end{align*}
Note that the kernel matrix produced by RBF Kernel is positive definite, however, kernel matrix produced by INK-spline of order zero is positive semidefinite.  
In the following we show that there is no solution exists for the XOR problem using v-SVM with INK-spline of order zero .

The Python code for fitting v-SVM model is given below.
\begin{lstlisting}
# Import required library
import numpy as np
from numpy import linalg as LA
from cvxopt import matrix, solvers

def vSVM(X, y, kernel_method = rbf_kernel, V=None):
    """Fit the v-SVM model according to the given training data.
            """
    n_samples, n_features = X.shape
    y = y.reshape(n_samples, 1)

    I_n = np.diag(np.ones(n_samples))  #Identity matrix
    if V is None:
        V = I_n  # Identity matrix

    # Compute the kernel matrix K 
    K = np.zeros((n_samples, n_samples))
    for i in range(n_samples):
        for j in range(n_samples):
            K[i, j] = kernel_method(X[i, :], X[j, :])

    KV = np.matmul(K, V)
    P = np.matmul(KV, K) + C * K
    P = matrix(P, (n_samples, n_samples), tc="d")
    q = np.matmul(KV, y)
    q = matrix(-q, (n_samples, 1), tc="d")
    A = matrix(np.matmul(K, np.ones(n_samples)), 
    		  (1, n_samples), tc="d")
    b = matrix(sum(y), tc="d")
    G1 = matrix(-K, (n_samples, n_samples), tc="d")
    h1 = matrix(np.zeros(n_samples), (n_samples, 1), tc="d")
    G2 = matrix(K, (n_samples, n_samples), tc="d")
    h2 = matrix(np.ones(n_samples), (n_samples, 1), tc="d")
    G = np.row_stack((G1, G2))
    h = np.row_stack((h1, h2))
    G = matrix(G, (2 * n_samples, n_samples), tc="d")
    h = matrix(h, (2 * n_samples, 1), tc="d")
    # solve the QP problem
    sol = solvers.qp(P, q, G, h, A=A, b=b)
    # Lagrange multipliers
    alpha = np.array(sol['x'])
    return alpha
\end{lstlisting}
The kernel functions are defined as following.
\begin{lstlisting}
# Radial basis kernel
def rbf_kernel(x1, x2, param=1.0):
    return np.exp(-(LA.norm(x1 - x2) ** 2) * param * 0.5)

# INK-spline of order zero
def ink_spline0_kernel(x1, x2):
    temp = np.minimum(x1, x2)
    return np.sum(temp)
\end{lstlisting}
The dataset for XOR problem is defined as follows.
\begin{lstlisting}
X =  np.array([[0, 0], [1, 1],[0, 1], [1, 0]])
y = np.array([np.zeros(2), np.ones(2)])
\end{lstlisting}
When we fit XOR data using v-SVM with RBF kernel, algorithms returns the optimal values.
\begin{lstlisting}
fit1 = vSVM(X,y, kernel_method = rbf_kernel)
\end{lstlisting}
However while trying to fit with the INK-spline kernel, v-SVM algorithm fails.
\begin{lstlisting}
fit2 = vSVM(X, y, kernel_method=ink_spline0_kernel)
\end{lstlisting}

\section{Another Example with Gaussian Mixture}
In the following we consider the Gaussian mixture model, two different Gaussian distributions for two different classes. In this case the kernel matrix produced by RBF kernel is ill-conditioned, whereas the INK-spline (with order zero) kernel function gives positive definite kernel matrix.  For the farmer case the v-SVM algorithm fails, and for later case the v-SVM returns the optimal values.

\begin{lstlisting}
# define some constants
n1 = 100
n2 = 100
l = n1+n2
mu1 = 1
mu2 = 10
sigma1 = 2
sigma2 = 3
C = 1

def generateData():
    x1 = np.random.normal(mu1, sigma1, n1)
    x2 = np.random.normal(mu2, sigma2, n2)
    X = np.concatenate((x1, x2), axis = 0)
    X = X.reshape((l,1))
    y = np.array([np.zeros(n1), np.ones(n2)])
    return X, y

X,y = generateData()
vSVM(X, y, kernel_method = rbf_kernel)
vSVM(X, y, kernel_method = ink_spline0_kernel)
\end{lstlisting}

\section{Conclusion}
The V-matrix method for estimation of conditional probability function is sensitive to the kernel matrix $K$. When the kernel matrix is ill-conditioned the inequality constraints are inconsistent and no solution could be found. The same argument holds true for the V-matrix methods  for estimation of regression function and estimation of density ratio function.

  \bibliographystyle{elsarticle-num} 
  \bibliography{L2_SVM}

\end{document}